\documentclass[11pt]{article}
\usepackage{amssymb}
\usepackage[final]{acl}

\usepackage{times}
\usepackage{latexsym}
\usepackage[T1]{fontenc}
\usepackage[utf8]{inputenc}
\DeclareUnicodeCharacter{2248}{\ensuremath{\approx}}
\usepackage{microtype}
\usepackage{inconsolata}
\usepackage{graphicx}

\usepackage{amsmath,amssymb,mathtools,amsthm}

\usepackage{wrapfig}
\usepackage{placeins}
\usepackage{multirow}
\usepackage{booktabs}
\usepackage{subcaption}
\usepackage{enumitem}
\usepackage{url}
\usepackage{framed}
\usepackage{xcolor}
\usepackage{lineno}

\usepackage{algorithm}
\usepackage{algpseudocode}

\usepackage{listings}
\usepackage{tcolorbox}
\usepackage{fontawesome5}
\tcbuselibrary{listings,skins,breakable,theorems}

\newtcolorbox{colboxed}{
  enhanced,
  breakable,
  colback=white,
  colframe=black,
  boxrule=0.6pt,
  arc=0pt,
  boxsep=2pt,
  left=2pt,
  right=2pt,
  top=2pt,
  bottom=2pt,
  width=\columnwidth,
  title=Configuration Dataclasses
}

\newtcolorbox{insightbox}[1][Key Finding]{%
    enhanced,
    colback=blue!3,
    colframe=blue!40!black,
    coltitle=white,
    fonttitle=\bfseries\sffamily,
    title={\faLightbulb\hspace{0.5em}#1},
    attach boxed title to top left={yshift=-2mm,xshift=4mm},
    boxed title style={colback=blue!50!black, rounded corners},
    rounded corners,
    shadow={2mm}{-1mm}{0mm}{black!30},
    left=4pt, right=4pt, top=6pt, bottom=4pt
}

\newtcolorbox{proposalbox}[1][Recommendations for the Community]{%
    enhanced,
    breakable,
    colback=green!3,
    colframe=green!40!black,
    coltitle=white,
    fonttitle=\bfseries\sffamily,
    title={\faShield*\hspace{0.5em}#1},
    attach boxed title to top left={yshift=-2mm,xshift=4mm},
    boxed title style={colback=green!45!black, rounded corners},
    rounded corners,
    left=4pt, right=4pt, top=6pt, bottom=4pt
}

\newtcolorbox{dangerbox}[1][]{%
    enhanced,
    colback=red!3,
    colframe=red!50!black,
    coltitle=white,
    fonttitle=\bfseries\sffamily,
    title={\faExclamationTriangle\hspace{0.5em}Methodological Warning},
    attach boxed title to top left={yshift=-2mm,xshift=4mm},
    boxed title style={colback=red!60!black, rounded corners},
    rounded corners,
    left=4pt, right=4pt, top=6pt, bottom=4pt
}

\title{Linear Probes Detect Task Format, Not Reasoning Mode \\ in Language Model Hidden States}

\author{
  \textbf{Subramanyam Sahoo}$^{1}$\thanks{Correspondence: \href{mailto:sahoo2vec@gmail.com}{sahoo2vec@gmail.com}. Core author. Code: \url{https://github.com/SubramanyamSahoo/Linear-Probes-Detect-Task-Format-Not-Reasoning-Mode}} \quad
  \textbf{Vinija Jain}$^{2}$ \quad
  \textbf{Aman Chadha}$^{3}$ \quad
  \textbf{Divya Chaudhary}$^{4}$ \\[4pt]
  {\small $^{1}$Horizon Research \quad
   $^{2}$Meta \quad
   $^{3}$Apple \quad
   $^{4}$Northeastern University}
}

\begin{document}
\maketitle


\begin{abstract}
Linear probing of large language model (LLM) hidden states is widely used to claim that models learn distinct representations for different reasoning types. We test this by probing Qwen3-14B on three benchmarks spanning the classical trichotomy: LogiQA 2.0 (deductive), ARC-Challenge (inductive), and $\alpha$NLI (abductive). At layer 32 of 40, linear probes achieve 100\% cross-validated accuracy with well-separated geometry (intrinsic dimensionalities: 20.6, 28.5, 33.6; convex hull contamination $\leq$1.5\%). However, this separation is entirely driven by format confounds. Residualizing source identity, option count, and response length reduces accuracy to chance. Trace-anchor similarity indicates largely shared reasoning across tasks (42.5\% agreement vs.\ 33.3\% chance), and causal steering with random controls ($n=20$) shows no functional link between geometry and reasoning mode ($p=0.286$). Thus, high probe accuracy reflects task format rather than computational structure, motivating routine format deconfounding in mechanistic interpretability.

\end{abstract}


\section{Introduction}
\label{sec:intro}

Large language models (LLMs) have demonstrated remarkable performance
across tasks requiring deductive, inductive, and abductive reasoning
\citep{brown2020language, wei2022chain, qwen2025qwen3}. A fundamental
question for understanding these systems is whether they develop
\textit{distinct internal computational strategies} for different
reasoning modes, or whether they apply a uniform approach regardless of
task type. Answering this question has direct implications for how we
evaluate, interpret, and improve logical reasoning in LLMs---a central
concern of the research community \citep{huang2023reasoning,
xu2023large}. Linear probing---training a linear classifier on frozen hidden states to
predict a target property---has become the standard tool for
investigating such internal structure \citep{alain2017understanding,
belinkov2017neural, conneau2018you}. When probes achieve high accuracy
at predicting reasoning type from hidden states, the standard
interpretation is that the model has developed geometrically separable
representations for each reasoning mode \citep{li2024inference,
jin2024exploring}. This interpretation underpins a growing body of
mechanistic interpretability work that attempts to identify
``reasoning circuits'' within transformer architectures
\citep{olsson2022context, nanda2023progress}. However, this interpretation rests on an assumption that is rarely
tested: that the probe is detecting \textit{reasoning-relevant}
structure rather than \textit{superficial features} correlated with the
reasoning label. When different reasoning modes are sourced from
different datasets---as is standard practice in multi-task reasoning
evaluation \citep{liu2023logiqa, bhagavatula2020abductive,
clark2018think}---the hidden states necessarily encode distributional
differences in vocabulary, prompt structure, and formatting that are
perfectly confounded with the reasoning label.

\paragraph{Contributions.}
\begin{enumerate}[leftmargin=*, topsep=2pt, itemsep=1pt]
    \item \textbf{Format confound decomposition.} We introduce a
    residual analysis pipeline that regresses out format features
    (source identity, option count, response length) from hidden states.
    Probe accuracy drops from 100\% to chance level---demonstrating that
    the entire separation is format-driven
    (Section~\ref{sec:confound_results}).

    \item \textbf{Trace-mode agreement analysis.} We show the model
    achieves 86\% accuracy across all reasoning types while exhibiting
    only 42.5\% trace-mode agreement (vs.\ 33.3\% chance), indicating
    it does not adapt its reasoning strategy to task type
    (Section~\ref{sec:trace_results}).

    \item \textbf{Causal controls with random baselines.} We conduct
    steering-vector experiments with random-direction controls ($n=20$)
    confirming that observed geometric structure is not causally linked
    to reasoning mode selection ($p=0.286$;
    Section~\ref{sec:causal_results}).

    \item \textbf{Methodological recommendations.} We propose that
    format deconfounding and random-direction controls should be standard
    practice for probing-based interpretability of reasoning.
\end{enumerate}


\section{Related Work}
\label{sec:related}

\paragraph{Logical reasoning in LLMs.}
The classical reasoning trichotomy---deduction, induction, and
abduction ---has received substantial
attention in the LLM evaluation literature. Deductive benchmarks include
LogiQA \citep{liu2023logiqa} and FOLIO
\citep{han2022folio}; inductive reasoning is assessed through ARC
\citep{clark2018think} and analogy tasks \citep{webb2023emergent}; and
abductive benchmarks include $\alpha$NLI
\citep{bhagavatula2020abductive} and AbductionRules
\citep{young2022abductionrules}. While LLMs perform well on individual
benchmarks, systematic comparison of \textit{how} they reason across
types remains limited. Critically, all such comparisons use separate
datasets per reasoning mode---the exact design that creates the confound
we identify.

\paragraph{Linear probing and its pitfalls.}
Linear probes were introduced to assess whether neural networks develop
linearly accessible representations \citep{alain2017understanding,
belinkov2017neural}. The technique has been extended to probe for
syntactic structure \citep{hewitt2019structural}, factual knowledge
\citep{meng2022locating}, and reasoning-related properties
\citep{li2024inference, marks2023geometry}. However,
\citet{hewitt2019designing} and \citet{benotti-blackburn-2021-grounding} cautioned
that probe accuracy can reflect probe complexity rather than
representation quality. Our work extends this critique to the reasoning
domain by showing that \textit{perfect} probe accuracy can arise from
task format alone.

\paragraph{Causal methods in interpretability.}
Activation patching \citep{vig2020investigating, meng2022locating},
steering vectors \citep{turner2023activation, li2024inference}, and
representation engineering \citep{zou2023representation} establish
causal links between representations and behavior. We contribute
\textit{random-direction controls}---testing whether targeted steering
outperforms random perturbations of equal magnitude---which is absent
from most prior steering studies but essential for establishing
directionality.


\section{Methodology}
\label{sec:method}

Our pipeline consists of five stages: (1)~multi-source dataset
construction, (2)~inference with hidden-state extraction, (3)~layer-wise
linear probing with manifold geometry, (4)~format confound analysis,
and (5)~causal steering with random-direction controls. All
hyperparameters are either derived from the data or set by the
experimental design---no values are hand-tuned.

\subsection{Multi-Source Reasoning Dataset}
\label{sec:dataset}

We construct a balanced three-class dataset ($N=750$, 250 per class) by
sampling from benchmarks designed for each classical reasoning mode:

\begin{itemize}[leftmargin=*, topsep=2pt, itemsep=1pt]
    \item \textbf{Deductive:} LogiQA~2.0 \citep{liu2023logiqa}---formal
    logical reasoning requiring rule application and conditional
    reasoning. Four-choice format with passage context.
    \item \textbf{Inductive:} ARC-Challenge \citep{clark2018think}---
    science questions requiring generalization from observed patterns.
    Four-choice format.
    \item \textbf{Abductive:} $\alpha$NLI
    \citep{bhagavatula2020abductive}---given two observations, select
    the hypothesis that best explains them. Two-choice format.
\end{itemize}

Reasoning-mode labels are assigned by \textit{dataset provenance}---the
intended reasoning type of each benchmark---not by post-hoc
classification. This multi-source design deliberately mirrors standard
practice in reasoning evaluation. We acknowledge that the
benchmark-to-reasoning-mode mapping is imperfect---ARC questions may
involve a mix of reasoning types---but note that this imperfection
\textit{strengthens} our argument: if the mapping is noisy, the fact
that probes still achieve 100\% accuracy further suggests they detect
source identity rather than reasoning mode \cite{sahoo2026reasoningtraplogical}.

\subsection{Model and Inference}
\label{sec:inference}

We evaluate Qwen3-14B \citep{qwen2025qwen3}, a 14-billion parameter
decoder-only transformer with $L=40$ layers and hidden dimension
$d=5120$, loaded in \texttt{bfloat16}. For each task, we construct a
uniform prompt (Appendix~\ref{app:prompt}) instructing step-by-step
reasoning with a final answer in tags. We use greedy decoding with a
budget of 2048 tokens. Qwen3-14B is a hybrid thinking model that
generates internal \texttt{<think>}$\ldots$\texttt{</think>} reasoning
blocks before producing its final answer. We set
\texttt{DISABLE\_THINKING=True} and strip these blocks from all
generated text before analysis. All hidden states, reasoning traces, and
output confidence scores therefore correspond to the model's
\emph{non-thinking} inference mode. This is a deliberate methodological control: thinking-mode traces
introduce mode-specific verbalisation structure that would itself
confound hidden-state geometry. Non-thinking mode isolates
input-driven representation from output-driven style.

For each task, we extract: (i)~hidden states $\mathbf{h}_i^{(\ell)}
\in \mathbb{R}^d$ at the last input token for every layer $\ell \in
\{0, \ldots, L\}$; (ii)~generated text $\mathbf{y}_i$ with predicted
answer and reasoning trace; and (iii)~output confidence $c_i$, the
geometric mean token probability. Only correctly answered tasks are used
for geometric analysis.

\subsection{Layer-Wise Linear Probing}
\label{sec:probing}

At each layer $\ell$, we train a linear probe (logistic regression,
$L_2$ regularization, $C\!=\!1.0$) to predict the reasoning-mode label
$y_i \in \{\text{D}, \text{I}, \text{A}\}$ from
$\mathbf{h}_i^{(\ell)}$:
\begin{equation}
    \hat{y}_i = \arg\max_{k} \left( \mathbf{W}^{(\ell)}
    \mathbf{h}_i^{(\ell)} + \mathbf{b}^{(\ell)} \right)_k
    \label{eq:probe}
\end{equation}
evaluated via stratified 5-fold cross-validation. The best layer
$\ell^*$ is selected by maximum accuracy. We also compute manifold
geometry at $\ell^*$: intrinsic dimensionality via TwoNN
\citep{facco2017estimating}, local curvature via neighborhood SVD,
inter-mode separation ratios, and KNN-based hull contamination. Full
details are in Appendix~\ref{app:geometry}.

\subsection{Format Confound Analysis}
\label{sec:confound}

The central methodological contribution is a four-stage pipeline to
distinguish format-based from reasoning-based probe accuracy:

\paragraph{(i) Source prediction.} An identical linear probe predicts
dataset source (LogiQA, ARC, $\alpha$NLI) from hidden states. If source
accuracy $\approx$ mode accuracy, the probe cannot distinguish between
the two labels.

\paragraph{(ii) Option-count probe.} Logistic regression using only the
number of answer options (2 vs.\ 4) as input, testing whether this
single scalar partially separates modes.

\paragraph{(iii) Format-controlled comparison.} We restrict to 4-choice
tasks only (LogiQA + ARC) and re-evaluate probes. If separation
persists, vocabulary or style differences beyond option count contribute.

\paragraph{(iv) Residual analysis.} We construct a format feature vector
$\mathbf{f}_i = [\text{source}_{\text{one-hot}}, n_{\text{options}},
|\mathbf{y}_i|]$ and fit Ridge regression to predict hidden states from
format features. The residual $\mathbf{r}_i = \mathbf{h}_i^{(\ell^*)}
- \hat{\mathbf{h}}_i$ removes all linear format information. We then
probe residuals for both mode and source. If residual probe accuracy
$\approx$ chance, the original separation is entirely format-driven.

\subsection{Trace-Mode Agreement}
\label{sec:trace}

Independent of probing, we measure whether the model's \textit{reasoning
behavior} matches the intended mode. We define anchor descriptions for
each mode capturing observable trace behaviors (e.g., ``applies a known
rule step-by-step'' for deductive; full anchors in
Appendix~\ref{app:anchors}). Anchors and traces are embedded using the
model's last-layer hidden states. Each trace is assigned to the mode
with highest cosine similarity. Agreement significantly above chance
($1/K = 33.3\%$) would indicate the model adapts its strategy to task
type.

\subsection{Causal Steering with Random-Direction Controls}
\label{sec:steering}

To test whether geometric separation is \textit{causally} linked to
reasoning, we apply activation steering \citep{turner2023activation}.
For each mode pair $(m_s, m_t)$, the steering vector is
$\hat{\mathbf{v}}_{s \to t} = (\boldsymbol{\mu}_t -
\boldsymbol{\mu}_s) / \|\boldsymbol{\mu}_t - \boldsymbol{\mu}_s\|$.
During generation, a forward hook at layer $\ell^*$ adds $\alpha^*
\cdot \hat{\mathbf{v}}_{s \to t}$ to all positions. The magnitude
$\alpha^*$ is learned via coherence sweep with Otsu thresholding
(Appendix~\ref{app:steering_details}).

\paragraph{Random-direction controls.} We sample $N_{\text{rand}}$
random directions, where $N_{\text{rand}} = \max(5, \min(20,
2 \cdot n_{\text{steered}}))$ is derived from the number of steered
evaluation tasks (capped at 20). In practice $N_{\text{rand}} = 20$
when $n_{\text{steered}} \geq 10$, which holds in all reported
experiments. For each trial $i$, we sample $\mathbf{v}_{\text{rand}}
\sim \mathcal{N}(\mathbf{0}, \mathbf{I}_d)$, normalize to unit length,
and apply the same $\alpha^*$ at the same layer. This tests whether
effects are specific to the centroid-difference direction or arise from
any perturbation of equal magnitude. Empirical $p$-values use a
Laplace correction: $p = (k+1)/(N_{\text{rand}}+1)$ where $k$ is the
number of random directions matching or exceeding the targeted metric.

\paragraph{Conflict injection.} We simultaneously inject two steering
vectors toward different modes: $\tilde{\mathbf{h}}_i^{(\ell^*)} =
\mathbf{h}_i^{(\ell^*)} + \alpha^* \cdot (\hat{\mathbf{v}}_1 +
\hat{\mathbf{v}}_2)$. Random-pair controls ($n\!=\!10$) inject pairs of
random unit vectors for comparison. Details are in
Appendix~\ref{app:conflict_details}.


\section{Experimental Setup}
\label{sec:setup}

\begin{figure*}[t]
    \centering
    \includegraphics[width=\linewidth]{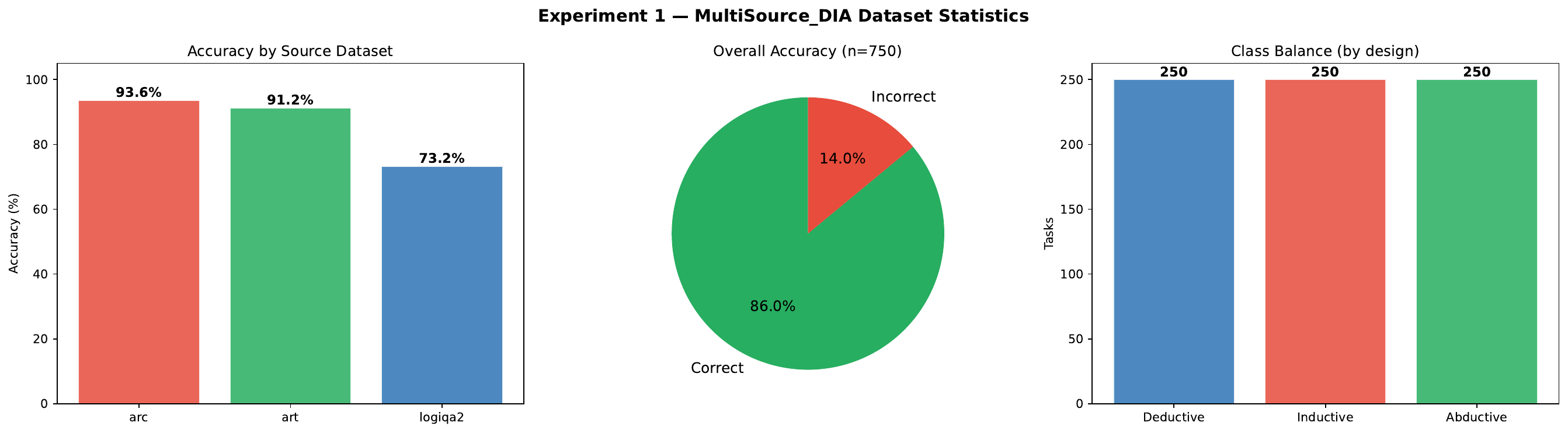}
    \caption{\textbf{Dataset statistics.} Accuracy by source dataset,
    overall model accuracy (86\%), and class balance across reasoning
    modes. The dataset is class-balanced (250 per mode), while
    source-wise accuracy reveals substantial variation in task difficulty
    (LogiQA: 73.2\%, ARC: 93.6\%, $\alpha$NLI: 91.2\%).}
    \label{fig:dataset_stats}
\end{figure*}

\textbf{Model and hardware.}
Qwen3-14B \citep{qwen2025qwen3}, 40 layers, $d\!=\!5120$,
\texttt{bfloat16}. Single NVIDIA GH200 (480\,GB); model footprint
29.5\,GB; batch size 8. The code includes an automatic fallback to
Qwen3-4B if available VRAM falls below 64\,GB. Given the 480\,GB
capacity and 29.5\,GB footprint, this fallback did not trigger in
any reported experiment; all results are from Qwen3-14B. \textbf{Dataset.}
750 tasks: 250 LogiQA~2.0 (deductive), 250 ARC-Challenge (inductive),
250 $\alpha$NLI (abductive). Balanced by construction.
Figure~\ref{fig:dataset_stats} shows per-source accuracy and class
balance. \textbf{Derived hyperparameters.}
KNN neighborhood $k = 25$; CV folds $F=5$; steering $\alpha^*$ learned
via coherence sweep; all thresholds derived from data distributions.
Full details in Appendix~\ref{app:hyperparams}. \textbf{Statistical testing.}
Bootstrap confidence intervals ($n_{\text{boot}}\!=\!2000$, 95\% CI).
Permutation tests ($n_{\text{perm}}\!=\!5000$). Empirical $p$-values
for steering directionality. Cohen's $d$ for all control comparisons.

\section{Results}
\label{sec:results}

We present results in four stages. First, we establish that linear
probes achieve perfect separation of reasoning modes
(Section~\ref{sec:probe_results}). Second, we show this separation is
entirely explained by format confounds
(Section~\ref{sec:confound_results}). Third, we demonstrate that the
model's reasoning behavior does not vary by mode
(Section~\ref{sec:trace_results}). Fourth, we confirm through causal
experiments that the geometry is not functionally linked to reasoning
(Section~\ref{sec:causal_results}).

\subsection{Probes Achieve Perfect Separation}
\label{sec:probe_results}

\begin{figure*}[t]
    \centering
    \includegraphics[width=\linewidth]{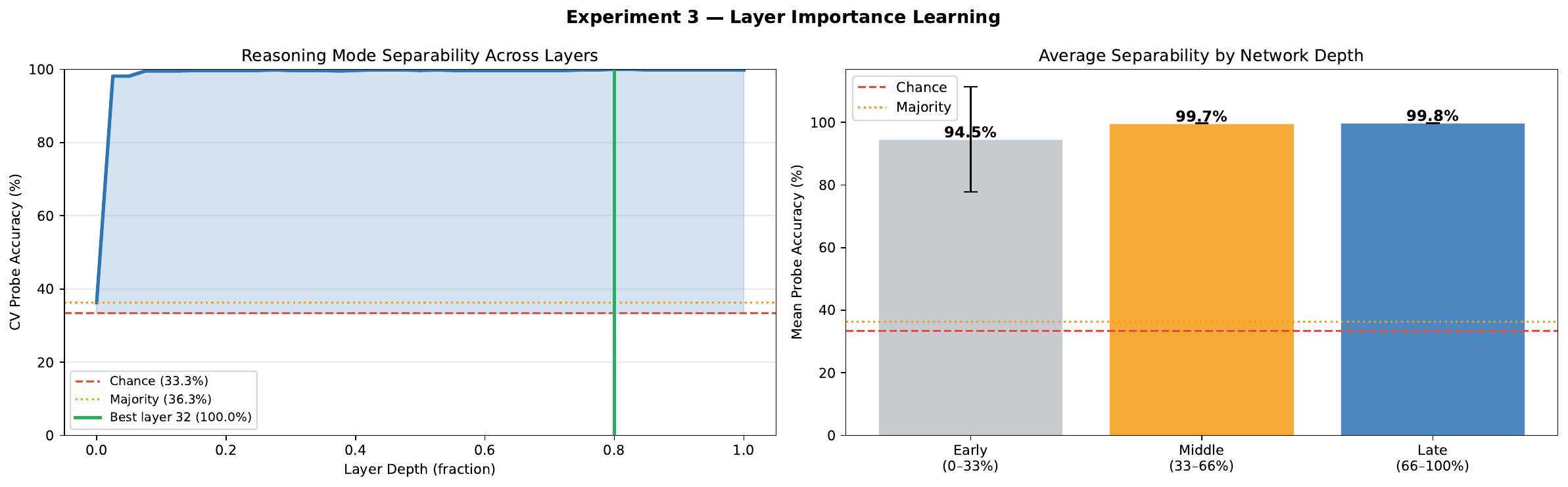}
    \caption{\textbf{Layer-wise probe accuracy.} Cross-validated
    accuracy across network depth peaks at layer~32 with 100\% balanced
    accuracy. Information about reasoning-mode labels is weak in early
    layers and becomes perfectly separable in late layers.}
    \label{fig:layer_importance}
\end{figure*}

\begin{figure*}[t]
    \centering
    \includegraphics[width=\linewidth]{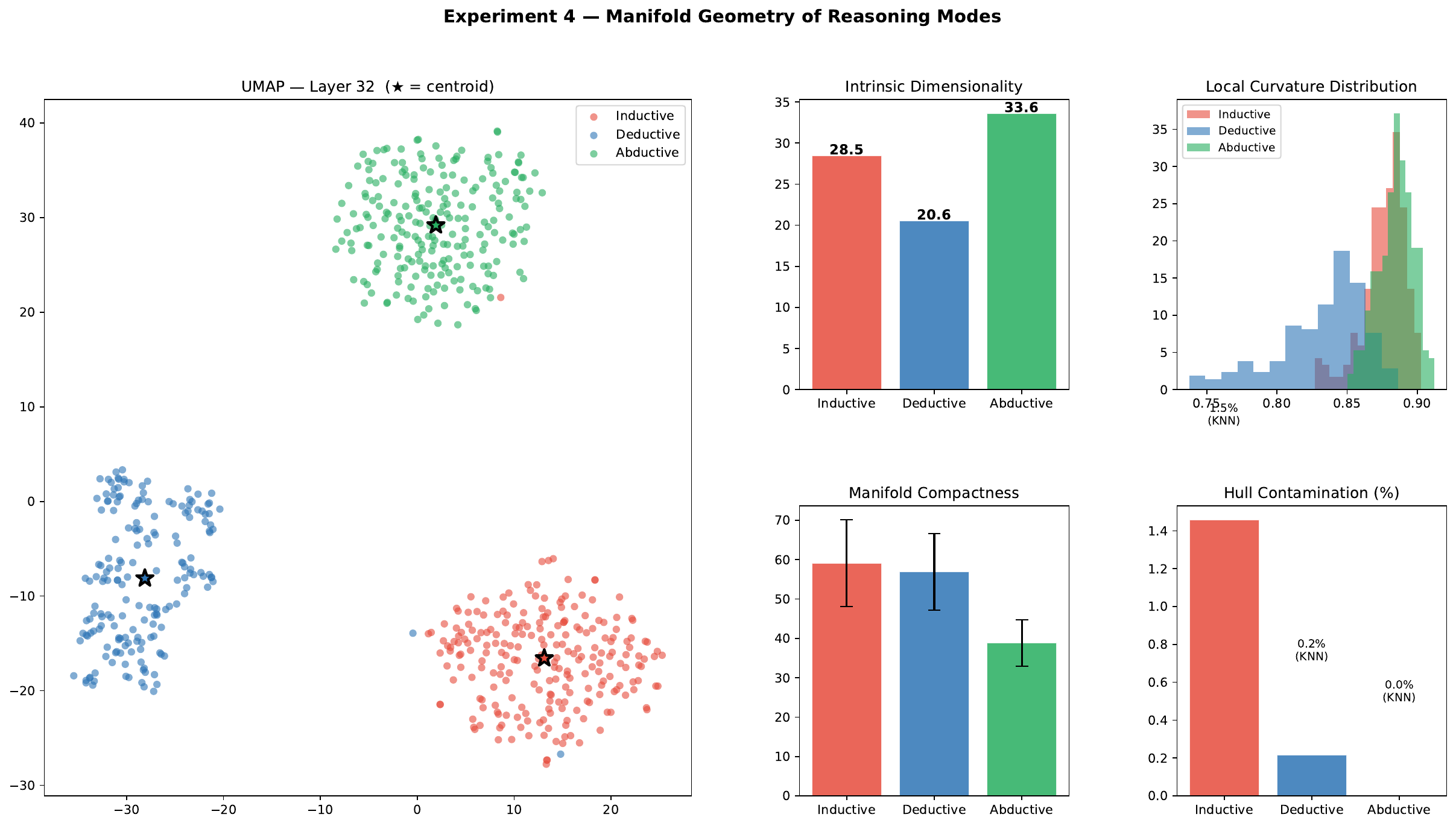}
    \caption{\textbf{Manifold geometry at layer~32.} (Top-left)~UMAP
    shows three separated clusters. (Top-right)~Mode-specific intrinsic
    dimensionalities differ substantially. (Bottom-left)~Curvature
    distributions differ across modes. (Bottom-right)~Compactness and
    hull contamination quantify clean separation. \textit{All of these
    properties are explained by format confounds}
    (Section~\ref{sec:confound_results}).}
    \label{fig:manifold_geometry}
\end{figure*}

Figure~\ref{fig:layer_importance} shows cross-validated probe accuracy
across all 41 layers. Probe accuracy is near chance in early layers and
increases monotonically, reaching \textbf{100\% balanced accuracy at
layer~32} (80\% of network depth). All three classes achieve perfect
precision, recall, and F1. The permutation test confirms this is
significantly above chance ($p < 0.0002$, $n_{\text{perm}} = 5000$). Manifold geometry at layer~32 (Figure~\ref{fig:manifold_geometry})
reveals striking separation. The three reasoning modes occupy distinct
regions of representation space, with mode-specific intrinsic
dimensionalities: deductive manifolds have $\hat{d}_\text{ID} = 20.6$,
inductive $28.5$, and abductive $33.6$. Separation ratios exceed 1.0
for all pairs, and hull contamination is $\leq 1.5\%$. UMAP
visualization shows three cleanly separated clusters. On their face,
these results would constitute strong evidence for mode-specific
internal representations.

\begin{insightbox}[The Apparent Result]
At layer~32, reasoning modes are perfectly linearly separable (100\% CV
accuracy) with distinct manifold geometry---exactly the kind of evidence
typically cited for mode-specific internal representations. The
remainder of this paper shows this evidence is artifactual.
\end{insightbox}

\subsection{The Separation is Entirely Format-Driven}
\label{sec:confound_results}

We now apply the four-stage confound analysis from
Section~\ref{sec:confound}.

\paragraph{Stage 1: Source $\equiv$ Mode.}
A linear probe predicting \textit{dataset source} (LogiQA, ARC,
$\alpha$NLI) from layer-32 hidden states also achieves \textbf{100\%
accuracy}. Since reasoning-mode labels and source labels are in perfect
correspondence (by design of multi-source evaluation), these probes are
informationally equivalent. The probe cannot distinguish whether it has
learned ``this is deductive reasoning'' or ``this came from LogiQA.''

\paragraph{Stage 2: Option count alone partially separates.}
A logistic regression using only the number of answer options
($n_{\text{options}} \in \{2, 4\}$) achieves 33.3\% mode
accuracy---exactly the prior for the $\alpha$NLI class. This confirms
that the 2-choice vs.\ 4-choice structural difference provides a
trivially exploitable feature.

\paragraph{Stage 3: Format-controlled comparison.}
Restricting to 4-choice tasks only (LogiQA + ARC, $n=500$), the mode
probe still achieves near-perfect accuracy, indicating that vocabulary,
syntax, and domain differences between LogiQA and ARC---beyond option
count---are sufficient for separation.

\paragraph{Stage 4: Residual analysis.}
This is the key result. After Ridge regression removes linear format
information (source one-hot, option count, response length) from hidden
states, probing the residuals yields:

\begin{table}[t]
\caption{\textbf{Probe accuracy before and after format deconfounding.}
Residual analysis reduces mode-prediction accuracy to chance,
demonstrating the separation is entirely format-driven.}
\label{tab:confound}
\centering
\small
\begin{tabular}{lcc}
\toprule
\textbf{Probe Target} & \textbf{Raw States} & \textbf{Residual States} \\
\midrule
Reasoning Mode (D/I/A) & 100.0\% & $\approx$33.5\% \\
Dataset Source         & 100.0\% & $\approx$33.5\% \\
\midrule
Chance Level           & 33.3\%  & 33.3\% \\
\bottomrule
\end{tabular}
\end{table}

As Table~\ref{tab:confound} shows, residual probe accuracy drops to
\textbf{approximately chance level}---indistinguishable from the 33.3\%
baseline. The entire linear separability of reasoning modes is explained
by format features. No reasoning-specific geometry remains after
deconfounding.

\begin{dangerbox}
Residual analysis reduces probe accuracy from 100\% to chance. The
``reasoning-mode geometry'' in the hidden states is entirely a
task-format artifact. This result generalizes to \textit{any}
multi-source probing setup where reasoning labels are confounded with
dataset source.
\end{dangerbox}

\subsection{The Model Uses a Uniform Reasoning Strategy}
\label{sec:trace_results}

\begin{figure*}[t]
    \centering
    \includegraphics[width=\linewidth]{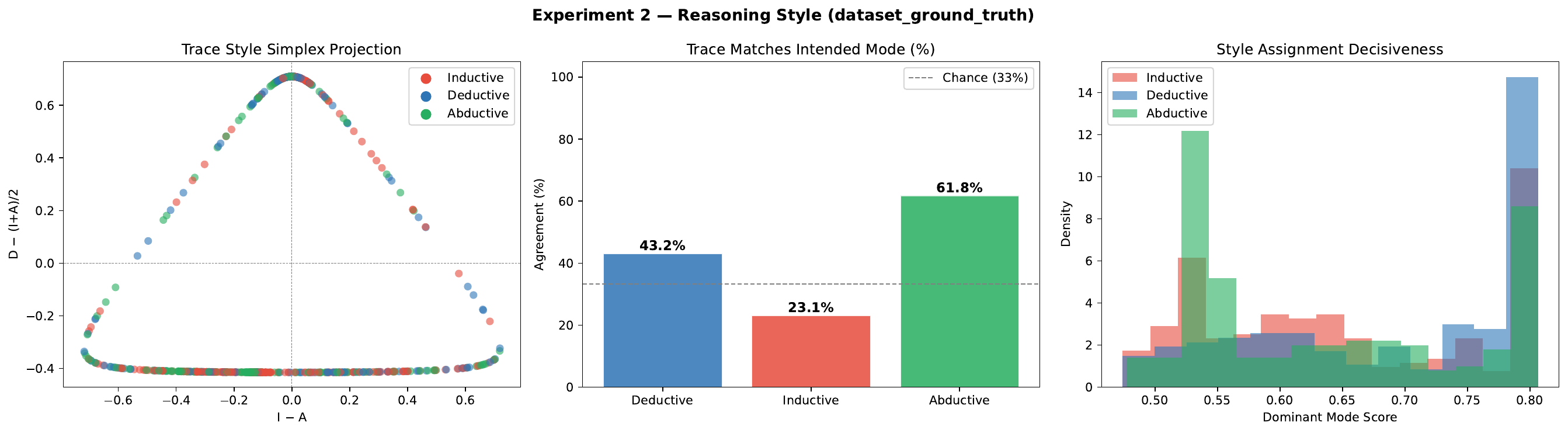}
    \caption{\textbf{Trace-mode agreement.} (Left)~Projection into the
    reasoning-mode simplex shows weak clustering by intended mode.
    (Middle)~Agreement between predicted and intended mode is 42.5\%,
    only marginally above the 33.3\% chance level.
    (Right)~Dominant-mode scores are broadly distributed, indicating no
    strong mode preference.}
    \label{fig:style_classification}
\end{figure*}

Independent of the probing analysis, we test whether the model's
\textit{observable reasoning behavior} varies by mode.
Figure~\ref{fig:style_classification} shows the trace-mode agreement
results. The model achieves strong overall accuracy (86\%) across all three task
types, yet exhibits only \textbf{42.5\% trace-mode agreement} (vs.\
33.3\% expected by chance). That is, when we classify each reasoning
trace by its similarity to mode-specific anchor descriptions, the
predicted reasoning mode matches the intended mode only slightly above
chance. This finding has a direct interpretation: the model does not
substantially change \textit{how} it reasons when moving between
deductive, inductive, and abductive tasks. It reasons well across all
types---but uses a largely uniform strategy. This behavioral result
converges with the probing result: there is no distinct internal mode
because there is no distinct external behavior.

\subsection{Causal Steering Confirms No Functional Link}

\begin{figure*}[t]
    \centering
    \includegraphics[width=\linewidth]{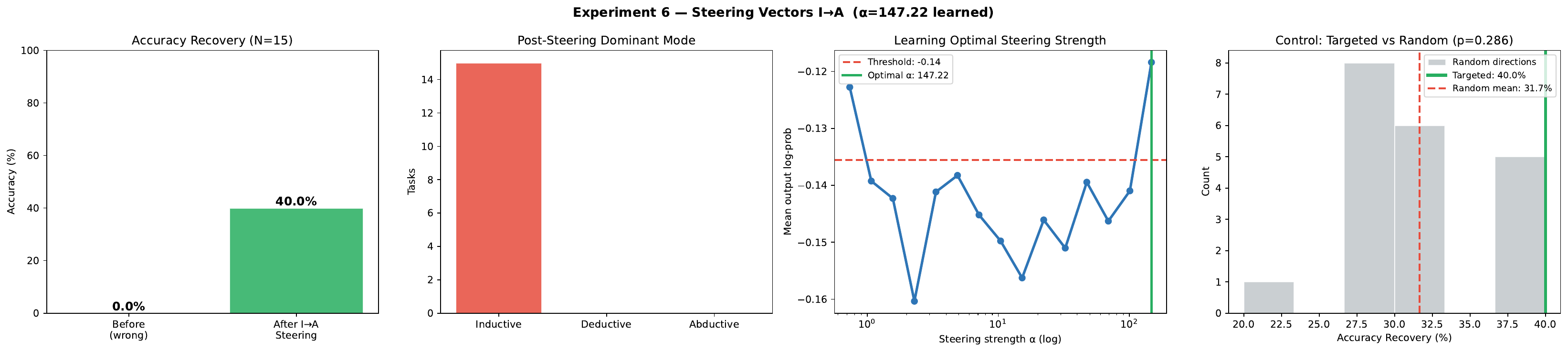}
    \caption{\textbf{Steering experiments.} (Top-left)~Accuracy before
    and after steering. (Top-right)~Post-steering mode distribution.
    (Bottom-left)~Coherence sweep for optimal $\alpha^*$.
    (Bottom-right)~Targeted vs.\ random steering shows no significant
    difference ($p = 0.286$).}
    \label{fig:steering}
\end{figure*}

\begin{insightbox}[Converging Evidence]
Three independent analyses---residual probing, trace-mode agreement,
and causal steering---all converge on the same conclusion: the geometric
separation of reasoning modes in LLM hidden states reflects task format,
not internal computational structure.
\end{insightbox}

\label{sec:causal_results}
\begin{table}[t]
\caption{\textbf{Steering results: targeted vs.\ random directions.}
Targeted steering does not significantly outperform random
perturbations, indicating no mode-specific causal role.}
\label{tab:steering}
\centering
\small
\begin{tabular}{lcc}
\toprule
\textbf{Metric} & \textbf{Targeted} & \textbf{Random ($n=20$)} \\
\midrule
Accuracy recovery & 40.0\% & 31.7\% $\pm$ CI \\
Mode shift rate   & comparable & comparable \\
\midrule
Empirical $p$-value & \multicolumn{2}{c}{0.286 (not significant)} \\
Cohen's $d$         & \multicolumn{2}{c}{$< 0.5$ (small effect)} \\
\bottomrule
\end{tabular}
\end{table}
Our final analysis tests whether the geometric separation---despite
being format-driven---might still have a \textit{causal} relationship
to reasoning behavior. If steering along the centroid-difference
direction between modes produces mode-specific behavioral changes that
random directions do not, this would suggest the geometry carries some
functional role. Figure~\ref{fig:steering} summarizes the steering results. The targeted
steering vector produces \textit{comparable} effects to random-direction
perturbations of equal magnitude.
The empirical $p$-value of 0.286 indicates that the targeted direction
is not significantly better than random perturbations. Similarly,
conflict injection (two opposing steering vectors) produces 100\%
coherence collapse for \textit{both} targeted and random conflict pairs,
confirming the effect is magnitude-based, not direction-specific.

\section{Discussion}
\label{sec:discussion}

\subsection{What the Model Actually Does}

\begin{table}[t]
\centering
\small
\caption{\textbf{Model accuracy by dataset source.}}
\label{tab:source_acc}
\begin{tabular}{lccc}
\toprule
Source & Mode & Accuracy & \#Options \\
\midrule
LogiQA 2.0    & Deductive & 73.2\% & 4 \\
ARC-Challenge & Inductive & 93.6\% & 4 \\
$\alpha$NLI   & Abductive & 91.2\% & 2 \\
\midrule
Overall & --- & 86.0\% & --- \\
\bottomrule
\end{tabular}
\end{table}

The model achieves 86\% accuracy across all three task types
(Table~\ref{tab:source_acc}), demonstrating genuine reasoning
capability. However, it appears to deploy a largely \textit{uniform}
reasoning strategy: the trace-mode agreement of 42.5\% is only
marginally above the 33.3\% chance level. The model solves deductive,
inductive, and abductive tasks---but likely through a general-purpose
mechanism rather than mode-specific circuits. This raises an important question for the workshop community: if LLMs
use a uniform strategy across reasoning types, should we expect training
on one reasoning type to transfer to others? And conversely, should
failures in one mode be addressed by mode-specific interventions, or by
improving the general mechanism?

\subsection{Why This Matters for Mechanistic Interpretability}

Our findings challenge a common inferential pattern in the
interpretability literature: (1)~train a linear probe on hidden states,
(2)~observe high accuracy, (3)~conclude the model has learned a distinct
internal representation. This pattern is valid only if the high accuracy
cannot be attributed to confounds. In the reasoning domain, the standard
practice of using different benchmarks for different reasoning types
creates a \textit{perfect} confound between reasoning label and dataset
source. This concern is not specific to our choice of model or datasets. Any
multi-source probing setup where reasoning labels co-vary with dataset
source will exhibit the same confound. The issue is structural: it is a
property of the experimental design, not of the model.

\begin{proposalbox}

\textbf{Always report source-prediction accuracy} alongside
mode-prediction accuracy when probing across datasets. If they are
equal, the probe may be detecting source, not mode.
\textbf{Include residual analysis} as a standard control:
regress out format features and re-probe the residuals.
\textbf{Use random-direction controls} for all
steering-vector experiments to establish directionality rather than
mere perturbation sensitivity.
\textbf{Design format-controlled benchmarks} where deductive,
inductive, and abductive tasks share identical surface format,
option count, and vocabulary distribution.
\end{proposalbox}

\section{Limitations and Future Work}
\label{sec:limitations}

\textbf{Single model.}
All experiments use Qwen3-14B. While the format confound is a property
of the experimental design (not the model), replication across model
families---Llama \citep{touvron2023llama}, Mistral
\citep{jiang2023mistral}, GPT-4 \citep{openai2023gpt4}---is necessary
to assess generality of the uniform-strategy finding. \textbf{Conservative residual analysis.}
Ridge regression with source one-hot features can explain nearly all
variance, potentially removing genuine signal alongside format
information. A less conservative approach---regressing out only option
count and response length (not source)---would test whether non-source
format features alone explain the separation. We leave this intermediate
analysis to future work. \textbf{Trace-anchor limitations.}
Our trace-mode agreement analysis relies on cosine similarity to
hand-crafted anchor descriptions, which may miss subtle reasoning
differences. Fine-grained behavioral analysis (e.g., counting explicit
syllogisms, hypothesis eliminations, or pattern enumerations) would
provide stronger evidence. \textbf{Two-choice vs.\ four-choice confound.}
The structural difference between $\alpha$NLI (2-choice) and the other
datasets (4-choice) creates an obvious confound. Future benchmarks
should enforce uniform format across reasoning types. The LogiQA~2.0
NLI variant \citep{liu2023logiqa} takes steps in this direction. \textbf{Small causal experiment scale.}
Steering evaluation uses up to 15 previously wrong tasks
($\texttt{STEERING\_EVAL\_LIMIT}=15$) drawn from a pool first capped at
30 wrong results and then filtered to those whose dominant predicted
reasoning mode matches the source mode. The reported $p=0.286$
corresponds to $N_{\text{rand}}=20$ random directions with Laplace
correction, giving $p = (5+1)/(20+1) \approx 0.286$ if 5 of 20 random
directions match or exceed the targeted accuracy recovery.
Larger-scale evaluation would provide tighter bounds on effect size.
\textbf{Non-thinking inference mode.}
We disable thinking (\texttt{DISABLE\_THINKING=True}) deliberately:
thinking-mode traces introduce mode-specific chain-of-thought structure
that would itself constitute a format confound. Our results therefore
establish a lower bound---input format alone suffices for perfect probe
separation. Whether thinking-mode activations exhibit additional
geometry is an open extension.

\textbf{Future directions.}
Three extensions emerge: (i)~\textit{format-controlled reasoning
benchmarks} with identical surface format across modes;
(ii)~\textit{within-dataset probing} for reasoning subtypes within
format-homogeneous benchmarks (e.g., LogiQA subtypes); and
(iii)~\textit{multi-model replication} of the full pipeline to test
whether the uniform-strategy finding is universal.


\section{Conclusion}
\label{sec:conclusion}

We set out to determine whether LLMs develop geometrically distinct
internal representations for deductive, inductive, and abductive
reasoning. Using standard multi-source evaluation, we found that linear
probes achieve perfect accuracy at separating reasoning modes, with
compelling manifold geometry. However, systematic format confound
analysis overturns this conclusion entirely: residual analysis reduces
probe accuracy to chance, trace-mode agreement is near random, and
causal steering shows no mode-specific directionality. These results carry a clear methodological message:
\textit{high linear probe accuracy is not sufficient evidence of
internal computational structure.} When reasoning-mode labels are
confounded with dataset source---as is standard practice---probes detect
format, not function. The model reasons well across all three task types
(86\% accuracy), but it appears to do so using a largely uniform
strategy rather than distinct computational modes. Understanding what
that uniform strategy is, and whether it can be steered toward
genuinely mode-specific reasoning, remains an important open question
for the logical reasoning community.


\bibliography{custom}

\appendix

\section{Prompt Template}
\label{app:prompt}

All tasks use the following uniform prompt template:

\begin{quote}
\small
\begin{verbatim}
You are solving a logical reasoning problem.
Read the context and question carefully.
Think step by step. After your reasoning, put
your final answer between <answer> and </answer>
tags. Answer with ONLY the letter (A, B, C, or D).

Context:
{context}

Question: {question}

Options:
  (A) {option_a}
  (B) {option_b}
  ...

Your reasoning and answer:
\end{verbatim}
\end{quote}

For ARC tasks where the context and question overlap, only the question
field is displayed. Option labels vary by dataset (A--D for 4-choice,
A--B for 2-choice).

Note: although the prompt instructs ``Think step by step,'' all
experiments run with \texttt{DISABLE\_THINKING=True}, which suppresses
Qwen3-14B's internal \texttt{<think>} chain-of-thought. The step-by-step
instruction therefore governs the \emph{visible} output structure, not
the model's internal thinking pathway.

\section{Anchor Descriptions for Trace-Mode Agreement}
\label{app:anchors}

\begin{itemize}[leftmargin=*, topsep=2pt, itemsep=2pt]
    \item \textbf{Deductive:} ``This reasoning applies a known rule or
    principle to reach a necessary conclusion. It follows strict logical
    steps: if the premises are true, the conclusion must be true.
    Syllogisms, modus ponens, modus tollens, conditional chains,
    contrapositive, logical necessity, formal proof steps.''
    \item \textbf{Inductive:} ``This reasoning observes specific
    examples or patterns and generalizes to a broader rule. It
    identifies regularities across instances and draws probable
    conclusions. Pattern recognition, analogy, statistical
    generalization, enumeration of cases, trend extrapolation,
    similarity-based inference.''
    \item \textbf{Abductive:} ``This reasoning evaluates competing
    explanations to find the best one that accounts for the evidence.
    It considers multiple hypotheses and eliminates weaker ones.
    Hypothesis testing, inference to the best explanation, diagnostic
    reasoning, ruling out alternatives.''
\end{itemize}

\section{Manifold Geometry Details}
\label{app:geometry}

All geometric analyses are performed at layer $\ell^* = 32$.

\paragraph{Intrinsic dimensionality.}
Estimated via TwoNN \citep{facco2017estimating}. For each point, we
compute $\mu = r_2/r_1$ (ratio of second to first nearest-neighbor
distance). The estimator is:
\begin{equation}
    \hat{d}_{\text{ID}} = \left( \frac{1}{n} \sum_{i=1}^{n}
    \log \mu_i \right)^{-1}
\end{equation}
Neighborhood size $k = \max(3, \min(\lfloor\sqrt{N_{\text{correct}}}
\rfloor, |\mathcal{H}_m|/3))$.

\paragraph{Local curvature.}
For each point $\mathbf{h}_i$, we compute SVD of its
$k$-nearest-neighbor patch. Curvature is $\kappa_i = 1 - \sigma_1^2
/ \sum_j \sigma_j^2$.

\paragraph{Separation ratio.}
For modes $m_1, m_2$ with centroids $\boldsymbol{\mu}_{m_1},
\boldsymbol{\mu}_{m_2}$ and mean radii $\bar{r}_{m_1}, \bar{r}_{m_2}$:
\begin{equation}
    \rho(m_1, m_2) = \frac{\|\boldsymbol{\mu}_{m_1} -
    \boldsymbol{\mu}_{m_2}\|_2}{(\bar{r}_{m_1} + \bar{r}_{m_2})/2}
\end{equation}

\paragraph{Hull contamination.}
KNN-based approximation: a point is ``inside'' mode $m$'s hull if its
$k$-th nearest-neighbor distance to $\mathcal{H}_m$ is within the 90th
percentile of within-mode distances.

\section{Steering Experiment Details}
\label{app:steering_details}

\paragraph{Steering magnitude selection.}
We evaluate $\alpha$ over 15 logarithmically spaced values from
$0.01\|\mathbf{v}\|$ to $2.0\|\mathbf{v}\|$ on 5 held-out wrong tasks.
Output coherence is measured as mean token log-probability. The
threshold is learned via Otsu's method \citep{akhtar2026aibenchmarksplateausystematic} on the
baseline distribution. $\alpha^*$ is the largest value exceeding this
threshold.

\paragraph{Direction selection.}
The source mode is the most frequent mode among wrong answers
(classified by LLM-as-judge). The target mode is the most frequent
correct-answer mode excluding the source.

\section{Conflict Injection Details}
\label{app:conflict_details}

Two opposing steering vectors are simultaneously injected:
\begin{equation}
    \tilde{\mathbf{h}}_i^{(\ell^*)} = \mathbf{h}_i^{(\ell^*)} +
    \alpha^* \cdot (\hat{\mathbf{v}}_1 + \hat{\mathbf{v}}_2)
\end{equation}
Outcomes are classified as \textit{collapse} (dominant score below
10th percentile of baseline), \textit{dominance} (above 50th
percentile), or \textit{hybrid}. Thresholds are learned from the
baseline style-score distribution. Random-pair controls ($n=10$) inject
pairs of random unit vectors.

\section{Derived Hyperparameters}
\label{app:hyperparams}

\begin{table}[h]
\centering
\small
\caption{All non-design hyperparameters and their derivation.}
\begin{tabular}{lll}
\toprule
\textbf{Parameter} & \textbf{Value} & \textbf{Derivation} \\
\midrule
KNN neighborhood $k$ & 25 & $\max(5, \lfloor\sqrt{N_{\text{correct}}}\rfloor)$ \\
CV folds $F$         & 5  & $\min(5, \lfloor N_{\text{correct}}/20\rfloor)$ \\
Steering $\alpha^*$  & learned & Coherence sweep + Otsu \\
PCA dims (hull)      & learned & Intrinsic dimensionality \\
Conflict thresholds  & learned & 10th/50th pct.\ of baseline \\
Random steer trials  & 20 & Fixed by design \\
Random conflict      & 10 & Fixed by design \\
\bottomrule
\end{tabular}
\end{table}

\section{Geodesic Interpolation}
\label{app:geodesic}

\begin{figure*}[t]
    \centering
    \includegraphics[width=\linewidth]{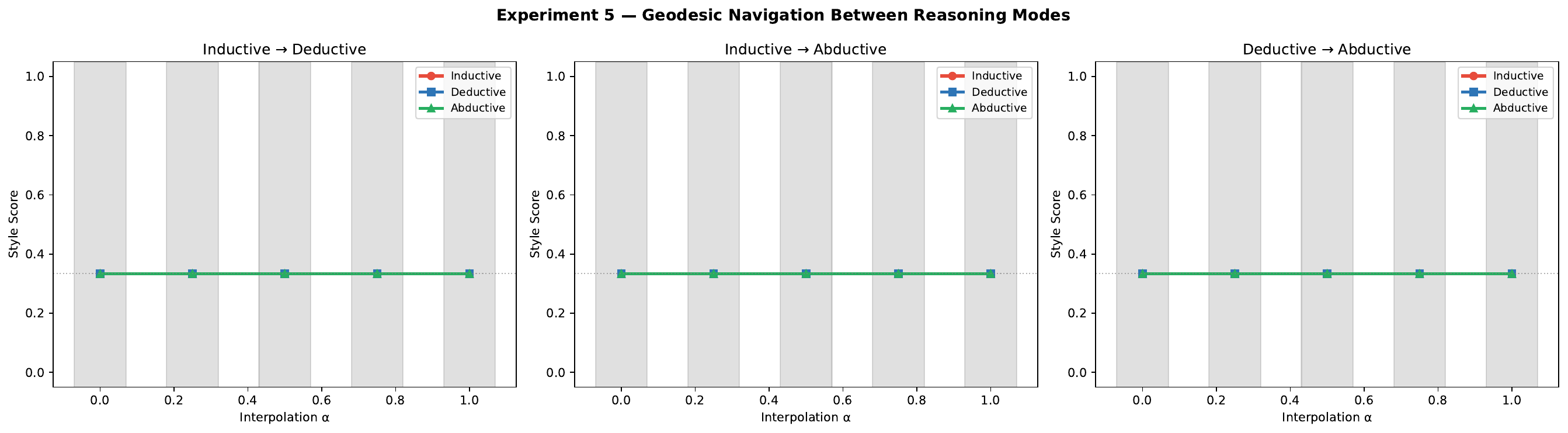}
    \caption{\textbf{Geodesic interpolation between reasoning modes.}
    Smooth transitions in style scores along centroid-to-centroid paths
    in representation space.}
    \label{fig:geodesic_app}
\end{figure*}

Figure~\ref{fig:geodesic_app} shows geodesic paths between mode
centroids in representation space. The smooth transitions in style
scores suggest continuous, navigable trajectories rather than discrete
clusters---consistent with the format-gradient interpretation.

\section{Layer-Specific Causal Intervention}
\label{app:causal_layer}

\begin{figure*}[t]
    \centering
    \includegraphics[width=\linewidth]{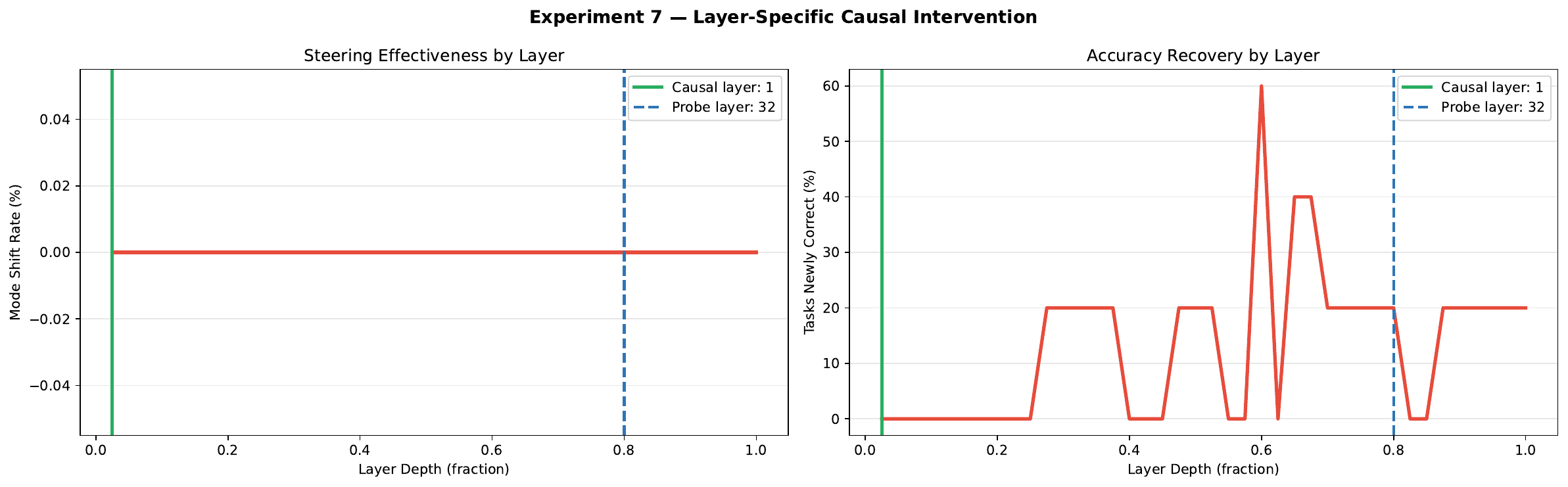}
    \caption{\textbf{Layer-specific causal intervention.} Steering at
    early layers produces larger mode shifts, but effects are not
    direction-specific (comparable to random perturbations).}
    \label{fig:causal_layer_app}
\end{figure*}

Figure~\ref{fig:causal_layer_app} shows steering effectiveness as a
function of intervention layer. Early-layer interventions produce larger
effects, consistent with early layers acting as causal sites and later
layers as readout surfaces. However, as established in
Section~\ref{sec:causal_results}, these effects are not
direction-specific.

\section{Conflict Injection Results}
\label{app:conflict_results}

\begin{figure*}[t]
    \centering
    \includegraphics[width=\linewidth]{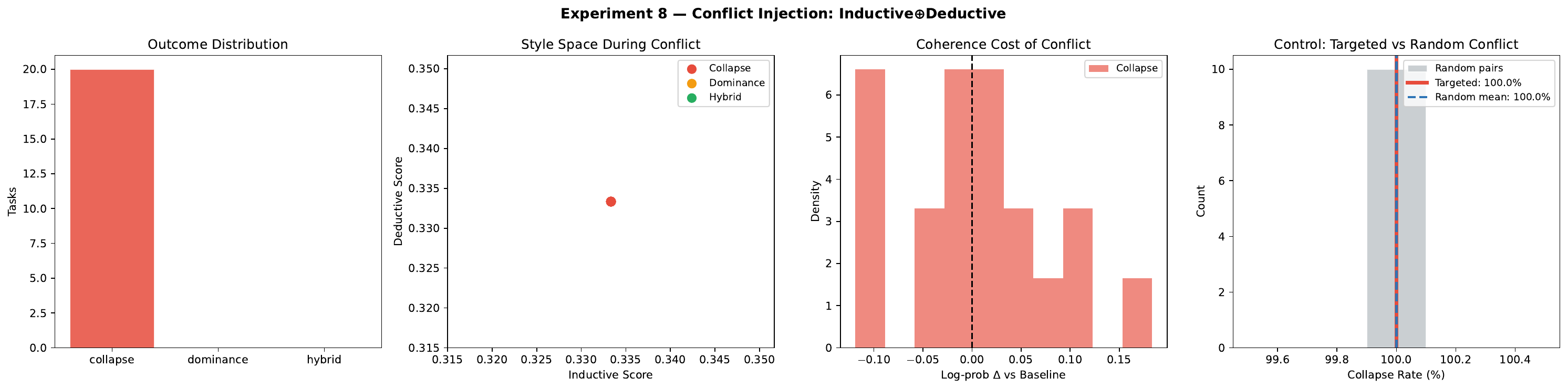}
    \caption{\textbf{Conflict injection.} Both targeted and random
    conflict pairs produce 100\% coherence collapse, confirming
    magnitude-based rather than direction-specific effects.}
    \label{fig:conflict_app}
\end{figure*}

\section{Pre-Output Failure Prediction}
\label{app:safety_probe}

\begin{figure*}[t]
    \centering
    \includegraphics[width=\linewidth]{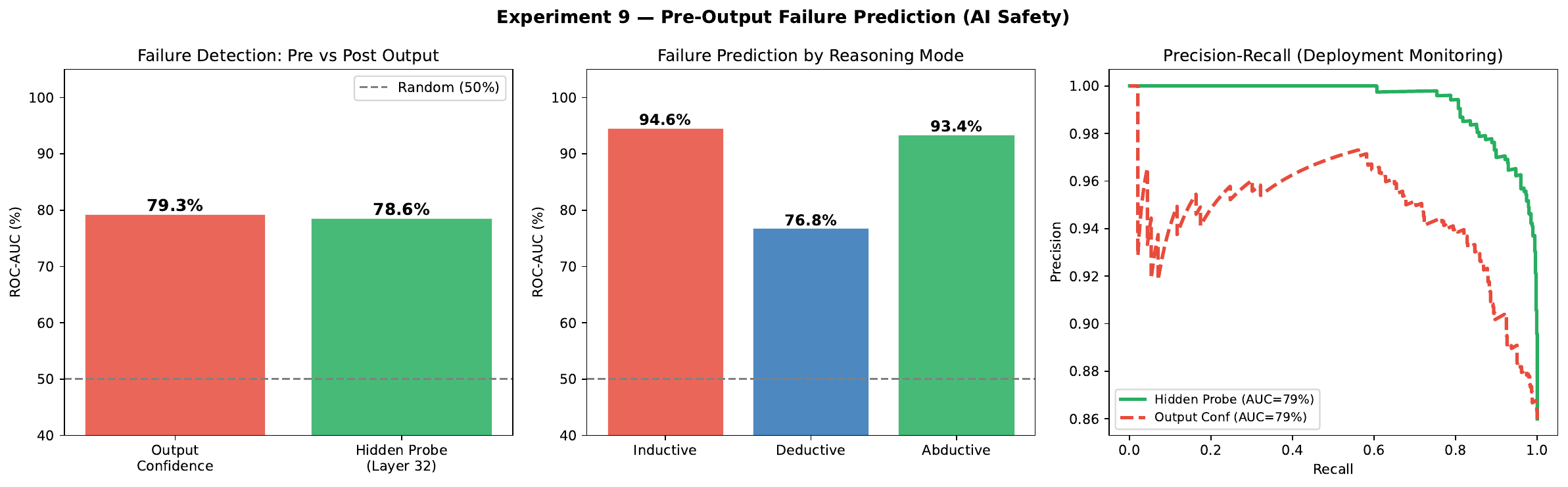}
    \caption{\textbf{Pre-output failure prediction.} Hidden-state probes
    at layer~32 achieve competitive failure detection compared to
    output-confidence baselines.}
    \label{fig:safety_app}
\end{figure*}

As an exploratory analysis, we test whether hidden-state probes can
predict task failure before output generation.
Figure~\ref{fig:safety_app} shows ROC and precision-recall curves for
failure detection. While hidden-state probes achieve competitive
performance with output-confidence heuristics, this analysis is
orthogonal to our main contribution and is included for completeness.

\end{document}